# Semantic Data Augmentation for Long-tailed Facial Expression Recognition

1st Zijian Li
School of Automation Science
and Electrical Engineering
Beihang University
Beijing, P.R.China
lizijian@buaa.edu.cn

2nd Yan Wang∗
School of Automation Science
and Electrical Engineering
Beihang University
Beijing, P.R.China
w-yan@buaa.edu.cn

3rd Bowen Guan
School of Automation Science
and Electrical Engineering
Beihang University
Beijing, P.R.China
861143011@qq.com

4th JianKai Yin
School of Automation Science
and Electrical Engineering
Beihang University
Beijing, P.R.China
SY2103130@buaa.edu.cn

*Abstract*—Facial Expression Recognition has a wide application prospect in social robotics, health care, driver fatigue monitoring, and many other practical scenarios. Automatic recognition of facial expressions has been extensively studied by the Computer Vision research society. But Facial Expression Recognition in real-world is still a challenging task, partially due to the long-tailed distribution of the dataset. Many recent studies use data augmentation for Long-Tailed Recognition tasks. In this paper, we propose a novel semantic augmentation method. By introducing randomness into the encoding of the source data in the latent space of VAE-GAN, new samples are generated. Then, for facial expression recognition in RAF-DB dataset, we use our augmentation method to balance the long-tailed distribution. Our method can be used in not only FER tasks, but also more diverse data-hungry scenarios.

*Keywords—Facial Expression Recognition, Data Augmentation, Long Tailed Recognition, VAE-GAN*

## I. INTRODUCTION

**Facial Expression Recognition.** Facial expressions are one of the most natural, ubiquitous and fundamental signals for humans to express emotional states and intentions. They could be used as powerful analysis cues in many scenarios. Facial Expression Recognition (FER) is the technology of identifying the expressions contained in the input emotion-related facial pictures. With the evolution of intelligent healthcare, smart-home, and man-machine interactive systems, FER has attracted more and more research interests. With the development of more and more powerful artificial neural networks, various works have successfully applied convolutional neural networks (CNN) and Attention Mechanisms to FER tasks[1-6]. EAC method utilizes filp semantic consistency of facial images to suppress noisy samples[1]. DeRL deposits expressions from mid-layers of a de-expression GAN[3]. DACL adaptively weights feature dimensions for center loss[4]. IF-GAN generates identity-free images for recognition[5].

Although in many lab-collected FER datasets, the recognition accuracy can be as high as 96.8% even several years ago[7], for many in-the-wild FER tasks, the state-of-the-art approaches are far more than satisfying. One key reason is that in-the-wild FER datasets, e.g. FER2013, Affectnet[8], SFEW[9], and RAF-DB[10], tend to be very imbalanced and suffer from the well-known Long-Tailed Recognition (LTR) problem. Fig. 1 shows the number of samples in each class of the RAF-DB dataset. Other datasets share similar distribution patterns.

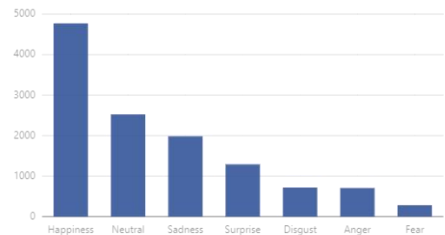

Fig. 1. Visualizing long-tailed data distribution of RAF-DB training set. Major classes and minor classes can be very imbalanced. There are 4772 samples with the label of Happiness while only 281 with the label of Fear.

In real-world applications, detecting abnormal expressions, such as anxiety and fear, is generally considered more important. But the difficulty of collecting abnormal expressions often results in a long-tailed distribution characteristic of FER tasks dataset. And as for currently dominant data-driven FER approaches, the long-tailed data leads to low recognition accuracy of rare expressions.

**Long-Tailed Recognition.** LTR problem is prevalent in many computer vision tasks. There are many existing methods to alleviate its influence. TADE aggregates inferences from diverse experts for test-agnostic Long-Tailed Recognition[11]. A recent study shows that weight balancing of output layers can alleviate the long tail problem[12]. Cost-Sensitive Weighting, Representation Learning, and many other techniques all find their spots in solving LTR problems[13]. But the most intuitive, widely used solution for LTR is trying to balance different categories of data. Over-sampling and under-sampling are probably the most common and easy-to-conduct options. However, when tail classes are too small, over/under-sampling may easily lead to severe over-fitting. Also aiming to balance the training data, some data augmentation techniques have been proven more effective.

**Data Augmentation.** Data augmentation generates new data from the original training datasets, and the generated data are used in the training process. It has been widely adopted in the field of computer vision tasks[14,15]. Traditional image augmentation methods like random cropping, rotation, or color jittering are not enough for LTR problems. Information aggregation is needed to enlarge tail classes. Linear interpolation between training samples or their extracted features can yields useful synthesized new data. SMOTE fuses same class samples into new ones[16]. And ReMix[17], CutMix[15], Mixaugment [18] and Mixup[19] fuse samples from different classes into new ones. All these methods achieve improvements in LTR tasks, but linearly combining samples in image space fails to enrich the training dataset at the semantic level.

Bengio et al. found that neural networks are capable of extracting disentangled features in the middle layers[20], and the features extracted tend to carry semantic information linearly. Many previous works have successfully exploited this characteristic. Deep Feature Interpolation (DFI) can edit facial images at semantic level. They map facial images to deep convolutional feature space, and move the mapped samples in certain directions[21], which introduce transformations like "making older" or "adding glasses". Center of wearing-glasses samples minus glasses-free samples can bring us the "adding glasses" direction. To find meaningful directions, positive and negative labeled samples are demanded, limiting its application in data augmentation tasks. ISDA randomly samples augmented directions by computed covariance of each class[22]. Metasaug extend ISDA into LTR scenarios[23]. Some other works also try to augment data in feature space[24]. M2m transfer sample from majority classes to minority classes[25]. But if only augment features extracted by neural networks but the original data, normally the feature extractor itself won't benefit from the augmentation, since it only receives non-augmented data while training.

Many other Data Augmentation approaches adopt generative models like Generative Adversarial Networks (GAN), Variational Auto-Encoder (VAE). Doping uses GAN to generate augmented data for language anomaly detection[26]. BAGAN[27] and IDA-GAN[28] are raised for image classification in LTR scenarios. Qin et al. use cycleGAN to modify common facial emotion data to get rare ones[29]. We further carry these methods step forward by implementing semantic data augmentation in our proposed VAE-GAN encoding space.

In our work, we first train a VAE-GAN to build the mapping between encoding space and image space. After the encoder maps data into the latent encoding space, we augment the encodings of source data. Then the augmented encodings are projected back into image space through the decoder. This gives new data that carry more semantic information. We test our proposed method in FER tasks. By generating tail class samples, the LTR problem of FER has been relatively solved.

## II. METHODOLOGY

### A. The VAE-GAN model and network training

**VAE-GAN networks.** The network for augmentation we used is a VAE-GAN model[30]. Our VAE-GAN consists of three sub-networks, i.e., an encoder $E$, a generator or decoder $G$, and a discriminator $D$. The whole model could be viewed as a VAE combined with a discriminator supervising the VAE output. The $D$ helps the VAE for high-fidelity generation and avoids blurring results. While in another aspect, this model can be viewed as a GAN and an encoder regularizing the GAN's input. The $E$ helps the GAN for training stability and lowers the risk of mode collapse. Fig. 2 illustrates the network structure.

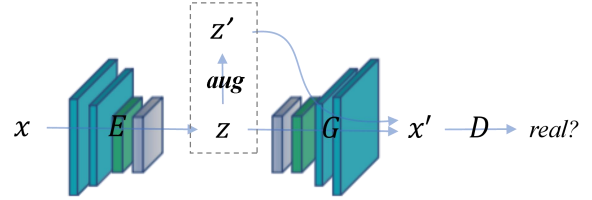

Fig. 2. VAE-GAN networks illustration. $E$, $G$, and $D$ are the three sub-networks introduced. $x$ is a training image sample and $x'$ is the networks' output correspondingly. The feature level augmentation, $z$ to $z'$, happens after VAE-GAN training.

In this paper, we use a Resnet[31] style encoder, while our generator and discriminator follow the design of BigGAN[32]. Encoder first maps an input $x$ into variance $E_{\sigma^2}(x)$ and mean $E_\mu(x)$ of a Gaussian distribution. Then randomly sample the encoding $z$ from that distribution. See (1), where $r$ is a random vector sampled from the standard Gaussian distribution, and the ∘ symbol stands for element-wise multiplication.

$$z = E_{\sigma^2}(x) \circ r + E_\mu(x) \sim \mathcal{N}(E_\mu(x), E_{\sigma^2}(x)) \qquad (1)$$

**Networks Training.** The losses used to train our networks include GAN losses, KLD loss $\mathcal{L}_{\mu,\sigma^2}$ and reconstruction loss $\mathcal{L}_r$. GAN losses follow the hinge loss that bigGAN uses. The KLD loss is the Kullback – Leibler divergence from $\mathcal{N}(E_\mu(x), E_{\sigma^2}(x))$ to $\mathcal{N}(0,1)$.

$$\mathcal{L}_{\mu,\sigma^2} = KL\left(\mathcal{N}(E_\mu(x), E_{\sigma^2}(x)) || \mathcal{N}(0,1)\right)$$
$$= \frac{1}{2}\left(E_\mu(x)^2 + E_{\sigma^2}(x) - \log(E_{\sigma^2}(x)) - 1\right) \qquad (2)$$

The reconstruction loss is the normal L1 distance of input $x$ and its reconstruction $x'$, plus perceptual reconstruction loss of weight $\lambda_p$[33].

$$\mathcal{L}_r = ||x - x'|| + \lambda_p ||P(x) - P(x')|| \qquad (3)$$

To avoid the KL vanishing problem[34], we linearly schedule the weight of $\mathcal{L}_{\mu,\sigma^2}$ from 0 to its maximum during the whole training process. Because the training of the VAE-GAN model is not the key point of this paper, we do not give detailed settings of this part.

### B. Semantic data augmentation

With the network properly trained, we can get the encoding of every sample. And our augmentation is done based on the latent encoding samples. In feature space, changing samples along certain directions can lead to certain semantic transformations, and this can be used for data augmentation. But sometimes such transformations may be meaningless, for example, adding tears and frowns to a sample from happiness class. So random sampling from isotropic distributions is not feasible. In our solution, for each class, the covariance matrix $\Sigma_c$ of all samples in class $c$ is computed. Then to augment a randomly selected sample $z$ in class $c$, we use (4) to add an augmentation random factor to the $E_\mu(x)$ part of $z$. In (4), $s$ is a predefined hyperparameter, controlling the augmentation strength.

$$z' = s\Sigma_c \circ r + E_\mu(x) \sim \mathcal{N}(E_\mu(x), s\Sigma_c) \quad (4)$$

The augmentation process of encoding is illustrated in Fig. 3, during which process, new semantic features are added to the original samples. Because we use covariance to decide the augmentation range, meaningless augmentation direction shall be omitted. After receiving the augmented feature $z'$, the decoder shall generate augmented images.

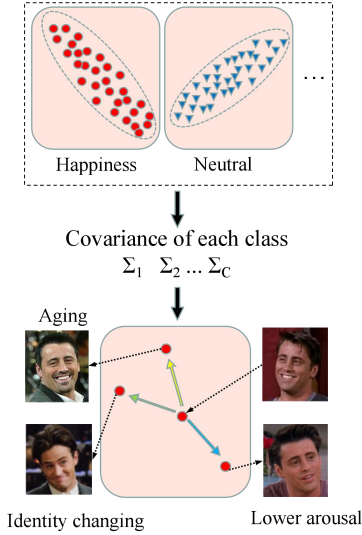

Fig. 3. Augmentation in the encoding space of VAE-GAN. This figure illustrates the process of $z$ to $z'$ in Fig. 2. First, the covariance $\Sigma_c$ of each class is computed. Then $\Sigma_c$ is used to sample the augmentation direction and step length. Lastly, original samples are modified accordingly. Different directions carry various semantic meanings in the latent space. Best viewed in color.

## III. EXPERIMENTAL RESULTS

**RAF-DB dataset.** There are 12,271 training and 3068 testing facial expression images in RAF-DB[35]. All images are downloaded from the Internet and annotated manually. Each image is cropped into 100×100 pixels and aligned properly. The training and testing sets both have similar long-tailed distributions.

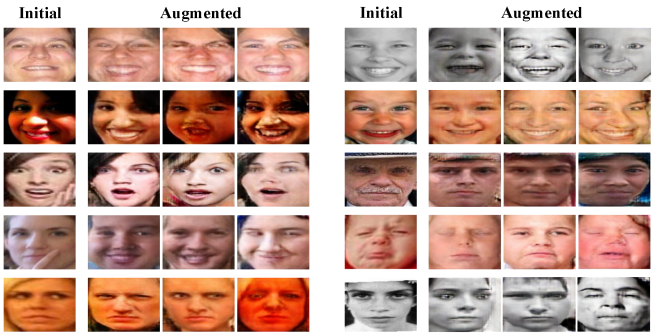

Fig. 4. Visualization of source images and their augmented ones. The augmentation process changes the posing direction, light, identity, and other high-level features of the data. Best viewed in color.

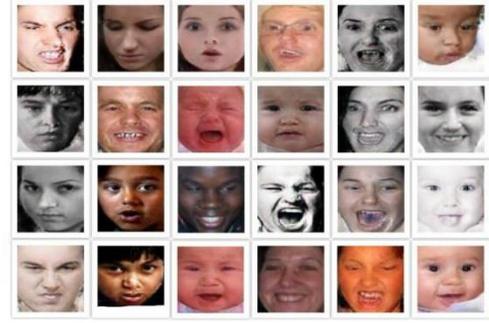

Fig. 5. Visualization of augmented images by our method. Classes are balanced through the augmentation. Best viewed in color.

**Visualization results.** Since our augmentation approach generates new data in image space, the most simple way to evaluate the proposed method is by plotting the augmented images and visually check them through human eyes. In Fig. 4, we can see that the augmentation process changes the posing direction, color, face shape, gender, age, and other high-level features of the data, implying that our method successfully augments data in semantic level. Fig. 5 shows more randomly selected augmented images. Our method enriches the dataset representations. And a diversified training dataset yields more robust networks.

**Numerical results.** For the imbalanced RAF-DB dataset, we use a balanced sampler to sample batches for training iteration, giving all classes equal frequency. By default, we set the augmentation ratio to 0.5, meaning half of the sampled training data will be augmented. We train classifiers 40 epochs. The training tick of only using the original training set in the last few epochs is adopted.

We test two classifiers. One is the classic Resnet34 model, the other is DAN[2], a leading model in FER tasks. DAN stands for "Distract Your Attention: Multi-head Cross Attention Network". It is composed of a Feature Clustering Network (FCN), a Multi-head cross Attention Network (MAN), and an Attention Fusion Network (AFN). FCN extracts features and clusters features from the same class together. Next MAN uses spatial attention and channel attention blocks to process features from FCN. Finally, AFN receives the attention maps and classifies the data.

TABLE I.    NUMERICAL RESULTS

| classifier | augmentation | Accuracy | |
|---|---|---|---|
| | | *Total Precision* | *mAP* |
| Resnet34[31] | None | 0.7021 | 0.5587 |
| Resnet34 | Balanced | 0.7322 | 0.6412 |
| Resnet34 | Ours | **0.7443** | **0.6737** |
| DAN[2] | None | 0.8832 | 0.8198 |
| DAN | Balanced | 0.8864 | 0.8095 |
| DAN | Ours | **0.8924** | **0.8205** |

Numerical results are reported in the table. Both total precision and mean averaged precision (mAP) are calculated.

Balanced means using resample to balance the long-tailed RAF-DB dataset. The experiment results prove the effectiveness of our augmentation method.

## IV. Conclusion

In this paper, we propose a novel data augmentation method that augments data at the semantic level. And experiments in FER tasks prove the effectiveness of our method. To our best knowledge, We are the first work to do semantic augmentation using class covariance, and maps between image and feature using a VAE-GAN model. Our next research is trying different augmentation rates for different classes in FER problems.

## Acknowledgment

This work is supported by the National Key Research and Development Project of China [Grant numbers 2018YFB2003501, 2020YFA0711200], and National Nature Science Foundation of China [Grant numbers 61320106010, 61573019, 61627810].